\documentclass[11pt,a4paper]{article}
\usepackage[hyperref]{emnlp2018}
\usepackage{times}
\usepackage{latexsym}

\usepackage{latexsym}
\usepackage{CJK}
\usepackage{url}
\usepackage{amsmath}
\usepackage{multirow}
\usepackage{booktabs}  
\usepackage{graphicx}
\usepackage{algorithm}  
\usepackage{algpseudocode} 
\usepackage{subfigure}
\usepackage{natbib}
\newcommand{\eqp}[1]{Eq.~\eqref{#1}}

\usepackage{url}

\aclfinalcopy % Uncomment this line for the final submission
\def\aclpaperid{1818} %  Enter the acl Paper ID here

\newcommand*{\email}[1]{\texttt{#1}}
\title{Diversity-Promoting GAN: A Cross-Entropy Based Generative Adversarial Network for Diversified Text Generation}

\author{
Jingjing Xu, \ 
Xuancheng Ren, \ 
Junyang Lin, \ 
Xu Sun \ 
\\ MOE Key Lab of Computational Linguistics, School of EECS, Peking University  \\
\email{\{jingjingxu,renxc,linjunyang,xusun\}@pku.edu.cn}}

\date{}

\begin{document}
\maketitle

%\sloppy

\begin{CJK}{UTF8}{gbsn}
\begin{abstract}  Existing text generation methods tend to produce repeated and ``boring'' expressions. To tackle this problem, we propose a new text generation model, called Diversity-Promoting Generative Adversarial Network (DP-GAN). The proposed model assigns low reward for repeatedly generated text and high reward for ``novel" and fluent text, encouraging the generator to produce diverse and informative text. Moreover, we propose a novel language-model based discriminator, which can better distinguish novel text from repeated text without the saturation problem compared with existing classifier-based discriminators. The experimental results on review generation and dialogue generation tasks demonstrate that our model can generate substantially more diverse and informative text than existing baselines.\footnote{The code is available at  \url{https://github.com/lancopku/DPGAN}}
\end{abstract}

\section{Introduction}

Text generation is an important task in Natural Language Processing (NLP) as it lays the foundation for many applications, such as dialogue generation, machine translation~\cite{DBLP:journals/corr/abs-1805-04871,unpaired-sentiment-translation}, text summarization~\cite{DBLP:journals/corr/abs-1805-04869}, and table summarization~\cite{DBLP:journals/corr/abs-1711-09724}. In these tasks, most of the systems are built upon the sequence-to-sequence paradigm \citep{DBLP:conf/nips/SutskeverVL14}, which is an end-to-end model that encodes a source sentence to a dense vector and then decodes the vector to a target sentence. The standard training method is based on Maximum Likelihood Estimation (MLE).

Although being widely applied, the conventional MLE training  causes systems to repeatedly generate ``boring" sentences, which usually are expressions with high frequency (e.g., ``I am sorry" in dialogue generation~\citep{DBLP:conf/naacl/LiGBGD16}). The major reason is that MLE encourages the model to overproduce high-frequency words.\footnote{ For example, the frequency ratios of ``the", ``and", ``was" are 4.2\%, 3.2\%, 1.5\% in real data, and they go up to 7.1\%, 4.6\%, 5.3\% in the MLE generated data on our review generation task.} The over-estimation of high-frequency words discourages the model from generating low-frequency but meaningful words in real data, which makes generated text tend to be repeated and ``boring''.

To tackle this problem, we propose a new model for diversified text generation, called DP-GAN. The key idea is to build a discriminator that is responsible for giving reward to the generator based on the novelty of generated text.  We consider the text that is frequently generated by the generator as the low-novelty text and the text that is uncommon in the generated data as the high-novelty text. Considering  most of the real-world sentences are novel and fluent, we treat the real-world text as the positive example and the generated text as the negative example to train the discriminator. Such training mechanism encourages the discriminator to give higher reward for the text that looks like real-world data.   The reward is fed back to the generator, which promotes the generator to generate diverse and fluent text via policy gradient. In this framework, a good discriminator that can assign reasonable reward for the generator is a critical component. 

%In adversarial learning, the discriminator is required to identify whether a sentence follows the generated data distribution. If a sentence is generated repeatedly, it can be easily identified by the discriminator and given low reward, while novel text usually receives high reward. Then, the reward is fed back to the generator, which encourages the generator to produce diverse text via policy gradient.

%For example, for a sentence $A$ with slightly low novelty and a sentence $B$  with extremely low novelty, the classifier gives them almost the same reward: $0.010$ and $0.011$. 

However, directly applying a classifier as the discriminator like most existing GAN models (e.g., SeqGAN~\citep{DBLP:conf/aaai/YuZWY17}) cannot achieve satisfactory performance. The main problem is that the reward given by the classifier cannot reflect the novelty of text accurately. First,  most existing classifier-based discriminators take the probability of a sequence being true as the reward. When a sentence fits the distribution of real-world text and is far from the generated data, the reward saturates and scarcely distinguishes the difference between these novel sentences. For example, for a sentence $A$ with mildly high novelty and a sentence $B$ with extremely high novelty, the classifier cannot tell the difference and gives them saturated reward: $0.997$ and $0.998$. Second, in our tasks, we find that a simple classifier can reach very high accuracy (almost $99\%$), which makes most generated text receive reward around zero because the discriminator can identify them with high confidence. It shows that the classifier also cannot distinguish the difference between low-novelty text. The reason for this problem is that the training objective of the classifier-based GAN is in fact minimizing the  Jensen-Shannon Divergence (JSD) between the distributions of the real data and the generated data~\citep{f-GAN}. If the accuracy of classifier is too high, JSD fails to measure the distance between the two distributions, and cannot give reasonable reward to the model for generating real and diverse text~\citep{DBLP:journals/corr/ArjovskyCB17}.
% and the generated data and  the real data can be easily distinguished with high confidence

%To encourage the generator to generate novel and fluent text like humans, we train the language model by giving the generated text low reward and the real-world text high reward. 

Instead of using a classifier, we propose a novel language-model based discriminator and use the output of the language model, cross-entropy, as the reward. The main advantage of our model lies in that the cross-entropy based reward for novel text is high and does not saturate, while the reward for text with low novelty is small but discriminative. The analysis of the experimental results shows that our discriminator can better distinguish novel text compared with traditional classifier-based discriminators.

Our contributions are listed as follows:
\begin{itemize}
\item We propose a new model, called DP-GAN, for diversified text generation, which assigns low reward for repeated text and high reward for novel and fluent text. 

\item  We propose a novel language-model based discriminator that can better distinguish novel text from repeated text without the saturation problem.

\item The experimental results on review generation and dialogue generation tasks show that our method can generate substantially more diverse and informative text than existing methods.
\end{itemize}

\section{Related Work}

%In this section, the related studies are introduced and we focus on the adversarial training and the text generation. 

%\subsection{Adversarial Learning}

A great deal of attention has been paid to developing data-driven
methods for natural language dialogue generation.
Conventional statistical approaches tend to rely extensively
on hand-crafted rules and templates, require interaction with
humans or simulated users to optimize parameters, or produce
conversation responses in an information retrieval fashion.
Such properties prevent training on the large corpora that are becoming increasingly available, or fail to produce novel natural language responses.

Currently, a popular model for text generation is the sequence-to-sequence model~\citep{DBLP:conf/nips/SutskeverVL14, DBLP:conf/emnlp/ChoMGBBSB14}. However, the sequence-to-sequence model tends to generate short, repetitive~\cite{DBLP:journals/corr/abs-1805-03989}, and dull text~\cite{jingjingxuemnlp18-02}. Recent researches have focused on developing methods to generate informative~\cite{jingjingxuemnlp18-01} and diverse text~\citep{DBLP:conf/emnlp/LiMSJRJ17, DBLP:conf/naacl/LiGBGD16, DBLP:journals/corr/abs-1709-08878, DBLP:conf/emnlp/ShaoGBGSK17}. Reinforcement learning is incorporated into the model of conversation generation to generate more human-like speeches~\citep{DBLP:conf/emnlp/LiMSJRJ17}. Moreover, there are also other methods to improve the diversity of the generated text by using mutual-information, prototype editing, and self attention~\citep{DBLP:conf/naacl/LiGBGD16, DBLP:journals/corr/abs-1709-08878, DBLP:conf/emnlp/ShaoGBGSK17}.

In this paper, to handle this problem, we propose to use adversarial training~\citep{DBLP:conf/nips/GoodfellowPMXWOCB14, DBLP:conf/nips/DentonCSF15, DBLP:conf/emnlp/LiMSJRJ17}, which has achieved success in image generation~\citep{DBLP:journals/corr/RadfordMC15, DBLP:conf/nips/ChenCDHSSA16, DBLP:conf/nips/GulrajaniAADC17, DBLP:journals/corr/BerthelotSM17}. However, training GAN is a non-trivial task and there are some previous researches that investigate methods to improve training performance, such as Wasserstein GAN (WGAN)~\citep{DBLP:journals/corr/ArjovskyCB17} and Energy-based GAN (EGAN)~\citep{DBLP:conf/nips/SalimansGZCRCC16, DBLP:conf/nips/GulrajaniAADC17, DBLP:journals/corr/ZhaoML16, DBLP:journals/corr/BerthelotSM17}.
GAN in text generation has not shown significant improvement as it has in computer vision. This is partially because text generation is a process of sampling in discrete space where the normal gradient descent solution is not available, which makes it difficult to train. There are some researches that focus on tackling this problem. SeqGAN~\citep{DBLP:conf/aaai/YuZWY17} incorporates the policy gradient into the model by treating the procedure of generation as a stochastic policy in reinforcement learning.~\citet{DBLP:journals/corr/RanzatoCAZ15} trains the sequence-to-sequence model with policy gradient for neural machine translation. ~\citet{DBLP:journals/corr/BahdanauBXGLPCB16} applies the actor-critic model on the same task.

\begin{figure}[t] 
\centering
\includegraphics[width = 0.9\linewidth]{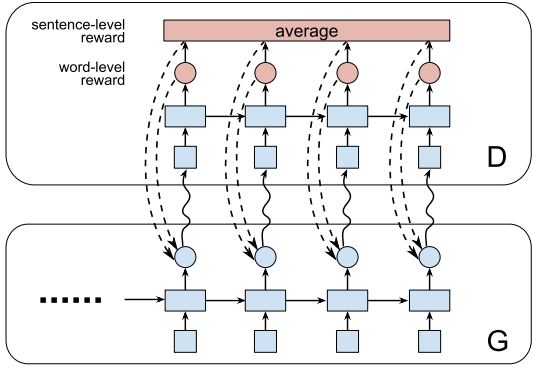}
\caption{Illustration of DP-GAN. Lower: The generator is trained by policy gradient where the reward is provided by the discriminator. Upper: The discriminator is based on the language model trained over the real text and the generated text.}
\label{dp-gan} 
\end{figure}

%Our PBGAN is a GAN consisting of a generator and a discriminator, which generates the texts and discriminates the generated data and real data respectively. In the following, an overview of the two parts is illustrated.
%****定义generator****

%Our generator is a standard sequence-to-sequence model, which consists of a RNN-based encoder as well as a RNN-based decoder. Both of them are built of Long Short-Term Memory network (LSTM) \citep{DBLP:journals/neco/HochreiterS97} (or GRU \citep{DBLP:conf/emnlp/ChoMGBBSB14})

\section{Diversity-Promoting GAN}

The basic structure of our DP-GAN contains a generator that is responsible for generating text and a discriminator that discriminates between the generated text and the real text. The sketch of DP-GAN is shown in Figure~\ref{dp-gan}.

\subsection{Overview}
The generator $G_{\theta}$ is based on a sequence-to-sequence structure. Given a sentence as input, the generator is capable of generating long text, which contains multiple sentences of various lengths. To put it formally, given the input sentence $x_{1:m} = (x_{1}, x_{2}, x_{3}, ..., x_{m})$ of $m$ words from $\Gamma$, the vocabulary of words,  the model generates the text of $T$ sentences $Y_{1:T} = (y_{1},...,y_{t},...,y_{T})$, where $y_{t}$ from $\Lambda$, the set of candidate sentence. The term $y_{t} = (y_{t,1},...,y_{t,K})$ is the $t^{th}$ sentence, where $y_{t,K}$ is the $K^{th}$ word.%is the $t^{th}$ sentence  $y_{t,k}\in\Gamma$. %$\Lambda$ and $\Gamma$ are the candidate sentence set and the vocabulary of candidate tokens. 

The discriminator $D_{\phi}$ is a language model. The output of the language model, cross entropy, is defined as the reward to train the generator. Our reward consists of two parts, the reward at the sentence level and that at the word level. With the discriminator and the reward function, we train the generator by reinforcement learning. A sketch of training DP-GAN is shown in Algorithm~\ref{code}. The details are described as follows.  %From this perspective, the generator is a policy, and the text generation is its trajectory with sampling a word at each time step as an action. Therefore, the reward at sentence level can be regarded as the final reward and the reward at word level can be regarded as the reward for each action, so the model can be trained with policy gradient.

%其次我们有一个$\phi$-parameterized discriminator $D_{\phi}$,$D_{\phi}$通过给generator反馈reward用来指导generator的学习。在我们的框架中，$D_{\phi}$是一个语言模型，reward为语言模型的困惑度。reward分为两种，一种是句子级别的reward，一种是单词级别的reward，句子级别的reward为$D_{\phi}(y_{t})$，通过综合考虑句子中所有单词的交叉熵来判定句子$y_{t}$是真文本的的概率，单词级别的reward为$D_{\phi}(y_t,k|t_{t, 1:k-1})$，也就是将每个单词的交叉熵作为reward返回给generator。

%****reinforcement learning****

%我们使用reinforcement learning的方法训练generaor.给定输入序列$X$, generator都可以产生一个输出序列，根据sampling from the policy $G_{\theta}$。对于generator而言，一方面discriminator会根据generator产生的样例给整个句子一个reward $D_{\phi}(y_{t})$，另一方面，discriminator还会给每个action一个reward$D_{\phi}(y_{t,k}|y_{t,1:k-1})$。接受到discriminator反馈的reward之后，$G_\theta$ 就可以按照policy gradient的方式进行参数更新。

%在$t$时刻，状态$s$是已经产生的tokens $(y_{1},...,y_{t-1})$, action是选择下一时刻的token $y_{t}$。 the policy model $G_{\theta}(y_t|Y_{1:t-1})$ is stochastic, whereas the state transition is deterministic after an action has been chosen。一般情况下，token $y_{t}$是根据$t$时刻，所有单词的出现概率sample得来。

%%%%%%%%%%%%%%%%%%%%%%%%%%%%%%%%%%%%%%%%%%%%%%%%%%%%
\begin{algorithm}[t]
\caption{The adversarial reinforcement learning algorithm for training the generator $G_{\theta}$ and the discriminator $D_{\phi}$.}
\label{code} 
\small
  \begin{algorithmic}[1]
%   \Require  
%      Random weights $\theta$, 
%       $\phi$,  
%       a dataset $\mathcal{D} = {(X,Y)}$ 
%    \Ensure
%        Generator $G_{\theta}$,
%       discriminator $D_{\phi}$  
      \State Initialize $G_{\theta}$, $D_{\phi}$ with random weights $\theta$, $\phi$ 
      \State Pre-train $G_{\theta}$ using MLE on a sequence dataset $\mathcal{D} = {(X,Y)}$
       \State Generate samples using $G_{\theta}$ for training $D_{\phi}$
       \State Pre-train $D_{\phi}$ by \eqp{eq:loss-d}
       \State $N$ = number of training iterations
        \State $M$ = number of training generator
         \State $K$ = number of training discriminator
       %\State $gstep$ = number of training generator;
       %\State $dstep$ = number of training discriminator;
     
    \For{each $i=1,2,...,N$}  
        \For{each $j=1,2,...,M$} 
        \State Generate a sequence $Y_{1:T}\sim G_{\theta}$
        \State Compute rewards by \eqp{eq:reward-sent} and \eqp{eq:reward-word}
        \State Update generator via policy gradient \eqp{eq:loss-grad}
        \State Sample a sequence $Y_{1:T}\sim \mathcal{D}$
        \State Compute rewards by \eqp{eq:reward-sent} and \eqp{eq:reward-word}
        \State Update generator parameters via \eqp{eq:loss-grad}
         \EndFor 
         \For{each $j=1,2,...,K$} 
         \State Generate samples using $G_{\theta}$
          \State Train discriminator $D_{\phi}$ by \eqp{eq:loss-d}
         \EndFor 
    \EndFor  
     
  \end{algorithmic} 
  
\end{algorithm}  
%%%%%%%%%%%%%%%%%%%%%%%%%%%%%%%%%%%%%%%%%%%%%%%%%%%%%%%%%%%%%%%%%%%%%%%%%%%%%%%

\subsection{Generator}

%为了模型的泛化性，本论文假设模型的输出可以是由任意句子组成的长的文本。%, where the encoder is a bidirectional LSTM and the decoder is a unidirectional LSTM
%Our generator is a RNN-based sequence-to-sequence model. In the sequence-to-sequence model, the encoder receives the representation of the source text, usually word embedding, and generates a compressed feature vector representing the whole source text, and the decoder receives the vector and generates the desired output. MLE is used in order to train the network. In the decoding of each word, attention mechanism \citep{DBLP:journals/corr/BahdanauCB14, DBLP:conf/emnlp/LuongPM15} is applied in our model so that the model can pay attention to the most relevant components in the source text, which improves the accuracy of the generation. 

For the concern of real-world applications, this paper assumes that the output of the model can be long text made up of multiple sentences. In order to generate multiple sentences, we build a standard hierarchical LSTM decoder~\citep{DBLP:conf/acl/LiLJ15}. The two layers of the LSTM are structured hierarchically. The bottom layer decodes the sentence representation and the top layer decodes each word  based on the output of the bottom layer. The attention mechanism is used for word decoding ~\citep{DBLP:journals/corr/BahdanauCB14, DBLP:conf/emnlp/LuongPM15}. %Moreover, we retrieve the last hidden state from the encoder as the context, which represents the information from the source text, and send it into the decoder at each time step of the decoding process. Therefore, when generating each sentence, the model can observe the last hidden state of the encoder, and thus can take the source information into consideration. %This method mitigates the vanishing problem in both the forward propagation and backpropagation, and encourages the decoder to generate longer sentences instead of ending up the generation.

% 介绍seq-seq

% encoder 部分将输入变成hidden-state

% 层次化的decoder，根据hidden state，先生成句子向量，再根据句子向量产生单词。其中产生单词的时候，用到了attention。

\subsection{Discriminator}
% 基于lstm的language model

Most existing GAN models use a binary classifier as the discriminator. The probability of being true is regarded as the reward~\citep{DBLP:conf/naacl/LiGBGD16,DBLP:conf/aaai/YuZWY17}. 
Different from that, we propose a language-model based discriminator $D_{\phi}$ that builds on a unidirectional LSTM. We use the output of the language model, cross-entropy, as the reward. Specifically, given a sentence $y_{t}$, the cross-entropy  based reward  for the $k^{th}$ word is calculated as 
\begin{equation*}
R(y_{t,k}) =- \log D_\phi ( y_{t,k}|y_{t,<k} )%, k = \{1,2,\ldots,K\}
\end{equation*}

We maximize the reward of real-world text and minimize the reward of generated text to train the discriminator. The reason of minimizing the reward of generated text is that, we expect the text that is repeatedly generated by the generator can be identified by the discriminator and  get lower reward. The motivation of maximizing the reward of real-world data lies in that, we expect not only the uncommon text in the generated data can get high reward, but also low-quality text can be punished to some extend. Considering the real-world text is diverse and fluent, we maximize the reward of real-world text to encourage the discriminator to give high reward for the text that looks like the real-world data. Therefore, such training mechanism avoids the problem of novel but low-quality text getting high reward. The loss function of the discriminator is formulated as follows:
\begin{align}
\begin{split}
\label{eq:loss-d}
&J(\phi) = \\
&- (E_{Y\sim p_{data}} [R(Y)]  - E_{Y\sim G_\theta} [R(Y)])
\end{split}
\end{align}
where $R(Y)$ stands for the averaged reward of $Y$. %To make the repeated generated text get lower reward, we minimize the expected reward of generated text. Furthermore, to make the novel and good-quality sentences get higher reward, we maximize the reward of the real data.  Therefore, those uncommon but low-quality text which does not follow the real data distribution will get lower reward, which can avoid the problem of learning from random sentences. 
%Furthermore, we maximize the expected reward of real data to make the language model learn the real data distribution, such that the generated sentences like diverse real data can easily get high rewards. By punishing repeated text and encouraging novel text, the generator is encouraged to generate diverse and informative expressions. 

%The discriminator is trained over the real data and the generated data. Minimizing the expected cross-entropy reward of generated data can be regarded as maximizing the output probability of generated data, such that the model is taught to fit the generated data distribution. After training, those repeatedly generated sentences can easily get high output probability and lower cross-entropy reward.

\subsection{Reward}
Our reward function consists of two parts, the sentence-level reward and the word-level reward, which are illustrated  as follows.

\subsubsection{Sentence-Level Reward}

For a sentence $y_{t}$ of $K$ words, the reward at the sentence level is the averaged reward of each word:
\begin{equation}
\label{eq:reward-sent}
R(y_t) = -\frac{1}{K}  \sum_{k=1}^{K} \log D_{\phi}(y_{t,k}|y_{t,<k})
\end{equation}

In contrast, the reward of the existing classifier-based discriminators~\citep{DBLP:conf/naacl/LiGBGD16,DBLP:conf/aaai/YuZWY17} is calculated as follows:

\begin{equation*}
R(y_t) = D_\phi(true|y_{t})
\end{equation*}
where $D_{\phi}$ is a binary classifier judging how likely $y_{t}$ is from the real-world data. 

The major problem of the classifier-based discriminator is that the reward cannot reflect the novelty of text accurately. First, the reward for high-novelty text is easy to saturate, which scarcely distinguishes the difference between novel text. Second, we find that the discriminator can easily achieve very high accuracy on identifying the generated text, which makes most of them get reward around zero. It shows that the classifier still cannot tell the difference between the text with low novelty.

On the contrary, the analysis of experimental result shows that our proposed discriminator can better distinguish high-novelty text from low-novelty text without the saturation problem. The reward for high-novelty text is high and does not saturate while the reward for low-novelty text is small but discriminative.%, so that it could encourage the model to generate diverse expressions. 

\subsubsection{Word-Level Reward}

%, rather than shares an averaged sentence-level reward,
Considering that the reward for different words in a sentence $y_{t}$ should be different, we further propose to use the reward at the word level as follows: 
\begin{equation}
\label{eq:reward-word}
R(y_{t,k}|y_{t,<k}) = - \log D_{\phi}(y_{t,k}|y_{t,<k})
\end{equation}
%where $D_{\phi}(y_{t,k}|y_{t,<k})$ is the probability of the $k^{th}$ word to be $y_{t,k}$.
It can be found that the classifier-based discriminator only provides reward for the finished sequence. Thus, for a sequence of length $T$, to evaluate the action-value for a word at the time step $t$,  Monte Carlo Search (MCS) with a roll-out policy $G_{\theta}$ is usually applied to sample the unknown last $T-t$ tokens~\citep{DBLP:conf/aaai/YuZWY17}.  However, this could be computationally expensive because the time complexity is $O(T^{2})$. On the contrary, our discriminator can calculate the reward of all words with the time complexity of $O(T)$, which is more computationally efficient.

%Besides, our model is capable of discriminating words in sentences due to the diverse cross entropies of the phrases of various novelty. 需要论据

%作为对比，在用classifier作为discriminator的传统的gan模型里面，因为分类器是根据整个句子进行打分的，为了区分不同单词的reward，一般会进行蒙特卡罗搜索。比如对t时刻的单词，一般会设定句子最大长度T，每次蒙特卡罗根据$G_{\theta}$ 按概率sample出T-t个单词得到一个完整的句子，共进行N次蒙特卡罗搜索，将N个句子得到的reward进行平均作为t时刻单词的reward。这样做的缺点在于，每个时刻t都要sample出N*(T-t)个单词，时间复杂度为$O(n^2)$。其次，每个时刻的reward还是根据完整的句子reward得来，在generator比较weak，classifier非常强的时候，几乎所有generated的句子都得到了接近于零的reward，无论怎么sample，每个时刻的reward几乎没有区分度。

%而我们的模型的优点在于只需要进行一次discriminator的推导就可以得出所有单词的reward，时间复杂度仅为$O(n)$，所以时间开销比较小。其次，新颖程度不同的短语的phrase的交叉熵是不同的，所以有能力对句子中的所有单词进行区分。

%As mentioned in the preceding section, if the generator is regarded as a stochastic policy, the generation of a text is a trajectory and the generation of each word is an action. Therefore, GAN can be trained by policy gradient with the cross entropy from the discriminator as the reward. 
\subsection{Policy Gradient Training}
The loss function of the generator (policy) is to maximize the reward from the start state $s_{0}$ to the end state~\citep{DBLP:conf/nips/SuttonMSM99}:
\begin{equation}
\label{equation1}
\begin{split}
J(\theta) & = \sum_{t=1}^{T} E[ R_{t,K}|s_{t-1},\theta] \\
           &= \sum_{t=1}^{T} \sum_{y_{t,1}} G_\theta (y_{t,1}|s_{t-1}) Q_{D_\phi}^{G_\theta}(s_{t-1}, y_{t,1})
\end{split}
\end{equation}
where $R_{t,K} = \sum_{k=1}^{K} \gamma^{k-1} R(y_{t})R(y_{t,k})$ is the total reward for a complete sentence, including both the sentence-level and the word-level rewards. The term $Q_{D_\phi}^{G_\theta}(s_{t-1}, y_{t,1})$ is estimated by $R_{t,1}$. The term $\gamma$ is the discount rate and $s_{t}$ is the initial state. 

In this paper, we use the policy gradient method~\citep{DBLP:journals/ml/Williams92}.
The gradient of~\eqp{equation1} is approximated as follows:
%using the likelihood ratio trick \citep{DBLP:journals/ml/Williams92, Glynn1990Likelihood, aleksandrov1968stochastic}:
%E_{Y_{1:T-1\sim G_\theta}}(\sum_{y_T\in \Gamma}G_{\theta}(y_t|Y_{1:t-1})Q_{D_{\phi}}^{G_{\theta}}(y_t,Y_{1:t-1})|\theta)
\begin{equation}
\label{eq:loss-grad}
\begin{split}
&\nabla_{\theta}{J(\theta)} \simeq  \\
&\sum_{t=1}^{T} 
%\frac{1}{K} 
\sum_{k=1}^{K} 
\gamma^{k-1} R_{t,k} \nabla_{\theta} \log G_\theta(y_{t,k}|y_{t,<k})
\end{split}
\end{equation}
where $R_{t,k} = \sum_{i=k}^{K} \gamma^{i-1} R(y_t) R(y_{t,i})$ is the total reward starting from step $k$.

%Considering that the generated data contains text with ungrammatical phrases, it is inevitable to give several low-quality text high reward. Therefore, to trade off fluency and diversity, we combine the MLE training method and the adversarial training together. 
Following previous work~\citep{DBLP:conf/emnlp/LiMSJRJ17}, we also use teacher forcing~\citep{DBLP:conf/nips/BengioVJS15} to train the generator. In teacher forcing, the decoder receives the real-world text as input at each time step. The loss function of teacher forcing is the same with that of policy gradient training. The only difference is that the text is generated from $G_{\theta}$ in policy gradient training but from the real data in teacher forcing. 
\section{Experiment}

We evaluate DP-GAN on two real-world natural language generation tasks, review generation and dialogue generation. We first introduce the dataset, the training details, the baselines, and the evaluation metrics. Then, we compare our model with the state-of-the-art models.  Finally, we show the experimental results and provide the detailed analysis.
\subsection{Datasets}

\textbf{Yelp Review Generation Dataset (Yelp)}:  This dataset is provided by Yelp Dataset Challenge.\footnote{\url{https://www.yelp.com/dataset/challenge}} In our version of review generation, the model should generate a paragraph based on a given sentence. We build a new dataset for this task by splitting the data into two parts. In each review, we take the first sentence as the input text, and the following sentences as the target text. The processed Yelp dataset contains 1,400K, 400K, and 12K pairs for training, validation, and testing, respectively. 

%
%The Amazon dataset is used for review generation. 
\noindent\textbf{Amazon Review Generation Dataset (Amazon):} This dataset is provided by~\citet{McAuley2013}. It consists of review information of fine foods from Amazon.  Like Yelp, we process this dataset by extracting the first sentence as the source text and the rest as the target text. The processed Amazon dataset contains 400K, 100K, and 12K pairs for training, validation, and testing, respectively. 

\noindent\textbf{OpenSubtitles Dialogue Dataset (Dialogue)}: This dataset\footnote{\url{http://opus.lingfil.uu.se/OpenSubtitles.php}} is used for dialogue generation. Following previous work, we treat each turn in the dataset as the target text and the two previous sentences as the source text. We remove the pairs whose response is shorter than 5 words. We randomly sample 1,800K, 500K, and 12K turns for training, validation, and testing, respectively.

%%%%%%%%%%

% \begin{table}[t]
% \centering
% \small
%     \begin{tabular}{l|l|c}
%     \hline

%     ~    &Model                & Averaged Ranking\\  \hline
%     \multirow{4}{*}{Yelp} & MLE   & 1.89    \\ 
%     &PG-BLEU           & 2.22       \\ 
%     &SeqGAN & 2.12 \\ 
%     &DP-GAN &  \textbf{1.51}\\ \hline
%      \multirow{4}{*}{Amazon} & MLE  & 1.93 \\ 
%     &PG-BLEU              & 2.24   \\ 
%     &SeqGAN &  1.98\\ 
%     &DP-GAN & \textbf{1.50}\\ \hline

%      \multirow{4}{*}{Dialogue} &MLE        &  2.46  \\ 
%    &  PG-BLEU & 2.40       \\
%     &  SeqGAN &  2.17        \\
%    & DP-GAN & \textbf{1.92}  \\ \hline
%     \end{tabular}
%     \caption{Results of human evaluation on three datasets.  Lower is better.}
%     \label{human}
%     %Ranking is evaluated in a comparative manner, where the annotator is shown a source sentence and the text generated by four models.
% \vspace{-1.0\baselineskip}
% \end{table}

\subsection{Baselines}

We compare the proposed DP-GAN with the following baseline models:

\noindent \textbf{MLE}: The generator is a sequence-to-sequence model. The generator is trained with traditional MLE. 

\noindent\textbf{PG-BLEU}: The generator is a  sequence-to-sequence model. It is trained by policy gradient with the BLEU score of the generated text as the reward~\citep{DBLP:journals/corr/BahdanauBXGLPCB16}. The advantage is that this model can directly optimize the task-specific score: BLEU.

\noindent\textbf{SeqGAN}: Sequence GAN~\citep{DBLP:conf/aaai/YuZWY17} uses a binary classifier as the discriminator. Since it is originally for unconditional generation, for a fair comparison, we expand it to the version of conditional generation. We re-implement the generator by replacing a language model with  a sequence-to-sequence model. %The discriminator, an rnn-based binary classifier, is required to evaluate sentences and give  rewards to guide the learning of the generator by reinforcement learning.

%%%%%%%
\begin{table}[t]
\footnotesize
\setlength{\tabcolsep}{3pt}
\centering
    \begin{tabular}{l|c|c|c|c|c}
    \hline
    Yelp & Token & Dist-1  & Dist-2 & Dist-3 & Dist-S\\ \hline
   MLE  &151.2K  & 1.2K & 3.9K & 6.6K & 3.9K             \\ 
    PG-BLEU &131.1K & 1.1K & 3.3K& 5.5K & 3.1K                    \\ 
    SeqGAN &140.5K  & 1.1K & 3.5K & 6.1K& 3.6K                \\ 
    \textbf{DP-GAN(S)} &\textbf{438.6K}    &1.7K &  7.5K & 15.7K &  10.6K               \\
     \textbf{DP-GAN(W)} &  271.9K & 2.8K & 14.8K  & 29.0K & 12.6K                \\
    \textbf{DP-GAN(SW) } & 406.8K & \textbf{3.4K} & \textbf{22.3K} & \textbf{49.6K} & \textbf{17.3K} \\ 
    \hline \hline
   Amazon & Token & Dist-1  & Dist-2 & Dist-3 & Dist-S\\ \hline
    MLE &   176.1K & 0.6K & 2.1K  & 3.5K  & 2.6K            \\ 
   PG-BLEU &   124.5K & 0.6K & 1.9K & 3.5K  & 2.3K           \\ 
    SeqGAN &     217.3K  & 0.7K  & 2.6K & 4.6K & 3.2K        \\ 
    \textbf{DP-GAN(S)} & \textbf{467.6K} &0.8K & 3.6K & 7.6K & 7.0K                \\
     \textbf{DP-GAN(W)} & 279.4K &1.6K & 8.9K & 18.4K& 9.6K               \\
    \textbf{DP-GAN(SW)} &   383.6K & \textbf{1.9K}  & \textbf{11.7K} & \textbf{26.3K} & \textbf{13.6K}               \\ \hline\hline
    
     Dialogue & Token & Dist-1  & Dist-2 & Dist-3 & Dist-S\\ \hline
     MLE  &  81.1K  & 1.4K & 4.4K& 6.3K  & 4.1K           \\ 
	PG-BLEU &   97.9K   & 1.2K & 3.9K & 5.5K  & 3.3K                \\ 
	SeqGAN &  83.4K    & 1.4K  & 4.5K  & 6.5K & 4.5K              \\ 
     \textbf{DP-GAN(S) } &  \textbf{112.2K}  & 1.5K & 5.2K  & 8.5K &   5.6K           \\
     \textbf{DP-GAN(W)} &  79.4K &  1.9K & 7.7K & 11.4K  &  6.0K             \\
	\textbf{DP-GAN(SW)}  & 97.3K &  \textbf{2.1K} & \textbf{10.8K}   & \textbf{19.1K}& \textbf{8.0K} \\ \hline

    \end{tabular}
    \caption{Performance of the DP-GAN and three baselines on review generation and dialogue generation tasks. Higher is better. DP-GAN(S), DP-GAN(W), and DP-GAN(SW) represent DP-GAN with only sentence-level reward, only word-level reward, and combined reward, respectively. \textsl{Token} represents the number of generated words. Dist-1, Dist-2, Dist-3, and Dist-S are respectively the number of distinct unigrams, bigrams, trigrms, and sentences in the generated text. For example, 1.2K in Dist-1 means 1200 distinct unigrams.  }
    \label{tab:auto-div}
\end{table}
% %%%%%%%%%%%%%%%%%%%%%%%%%%%%%%%%%%%%%%%%%%%%%%%%%%%%%%%%%%%%%%%%%

%%%%%%%%%%%%%%%%%%%%%%%%%%%%%%%%%%%%

\begin{table}[t]
\footnotesize
\setlength{\tabcolsep}{5.5pt}
\centering
    \begin{tabular}{l|c|c|c|c}
    \hline
    Yelp & Relevance & Diversity  & Fluency & All\\ \hline
    MLE  &    1.49 & 1.73 & 1.78& 1.89  \\ 
    PG-BLEU   &  1.47& 2.59 & \textbf{1.38}& 2.22\\
    SeqGAN &  1.48 & 2.40& 1.54&   2.12          \\

    \textbf{DP-GAN} & \textbf{1.32}& \textbf{1.23}& 1.66&\textbf{1.51}\\ 
    \hline \hline
   Amazon & Relevance &Diversity  &Fluency & All\\ \hline
    MLE  & 1.52 & 1.81 & 1.72 & 1.93      \\ 
     PG-BLEU &  1.62 & 2.48 &1.63& 2.24\\ 
    SeqGAN & 1.56& 2.37 &  \textbf{1.40} & 1.97           \\

    \textbf{DP-GAN} & \textbf{1.31} & \textbf{1.25}&1.52 &\textbf{1.50}\\ \hline \hline
     Dialogue & Relevance &Diversity  &Fluency & All\\ \hline
      MLE  &  1.19 & 1.84  &  1.37 & 1.87   \\ 
      PG-BLEU &  \textbf{1.13} &1.85 &1.21 &1.75\\ 
    SeqGAN & \textbf{1.13} &  1.71  &\textbf{1.20}  &   1.64      \\

    \textbf{DP-GAN} & \textbf{1.13}  &\textbf{1.50} &1.30 & \textbf{1.55}      \\ \hline

    \end{tabular}
    \caption{Results of human evaluation on the three datasets. The score represents the averaged ranking of each model and lower is better. \textsl{All} represents the ranking given by annotators based on a comprehensive consideration. It can be seen that DP-GAN results in the largest improvement in terms of diversity and relevance while slightly reducing fluency.}
    \label{human}
    %Ranking is evaluated in a comparative manner, where the annotator is shown a source sentence and the text generated by four models.

\end{table}

%%%%%%%%%%%%%%%%%%%%%%%%%%%%%%%%%%%%%%%%%%%%%%%%%%%%%%%%%%%%%%%%%%

\subsection{Training Details}
%Gradient clipping is adopted by scaling the gradients when their norms exceed a threshold of 2. 
%and the discount rate $\gamma$ is 0.5
For review generation, we set the number of generated sentences to 6 with the maximum length of 40 words for each generated sentence. Based on the performance on the validation set, we set the hidden size to 256, embedding size to 128, vocabulary size to 50K, and batch size to 64 for the proposed model and the baselines. We use the Adagrad~\citep{DBLP:journals/jmlr/DuchiHS11} optimizer with the initial learning rate 0.1. In adversarial training, the step for training the generator is 1K, the step for training the discriminator is 5K. Both the generator and the discriminator are pre-trained for 10 epochs before adversarial learning. In particular, for PG-BLEU and SeqGAN, before reinforcement learning or adversarial learning, we pre-train the sequence-to-sequence model for 10 epochs like DP-GAN. For dialogue generation, the settings are the same with review generation, except that we set the number of generated sentences to 1 with the maximum length of 40 words because there is only one sentence in the response.

\subsection{Experimental Results}

We conduct two kinds of evaluations in this work, automatic evaluation and human evaluation. The details of evaluation results are shown as follows.

\subsubsection{Automatic Evaluation}

%In review generation and dialogue generation tasks, it is widely debated how well the BLEU score against a single reference is correlated with the diversity of generated text~\citep{DBLP:conf/emnlp/LiuLSNCP16}. We suppose BLEU is not an adequate evaluation metric.

%Since we focus on improving the diversity of the generated text, 
%If a method generates more unique n-grams, it means the text covers more words or phrases, and is a reasonable indicator that the text is more diverse. 

We evaluate the proposed model in terms of several metrics that can reflect the diversity. The results are shown in Table~\ref{tab:auto-div}.~\textsl{Token} represents the total number of generated words. Dist-1, Dist-2, Dist-3, and Dist-S are respectively the number of distinct unigrams, bigrams, trigrms, and sentences. DP-GAN(S), DP-GAN(W), and DP-GAN(SW) represent DP-GAN with only sentence-level reward, only word-level reward, and combined reward, respectively. From the results, it is obvious that the proposed model substantially outperforms the existing models. PG-BLEU achieves slightly weaker results compared with MLE. The reason is that PG-BLEU uses BLEU score as the reward for reinforcement learning. However, the BLEU score is low for most of the generated text. The low reward makes it hard to learn from the real data. SeqGAN does not achieve better results, which suggests that the classifier-based discriminator fails to encourage the generator to produce diverse text.

In terms of the total number of generated words, DP-GAN(S) achieves better results than DP-GAN(W). Since the sentence-level reward reflects the novelty of the whole sentence, it  gives repeated and short text low reward while novel and longer text high reward. Thus, the generator is encouraged to generate novel text. In terms of the number of distinct n-grams, DP-GAN(W) achieves better results than DP-GAN(S). It is because the word-level reward gives each word more precise score and novel n-grams could be better encouraged. As we can see, DP-GAN(SW), which combines the advantages of sentence-level and word-level rewards, generates not only more diverse n-grams than DP-GAN(S) but also longer text than DP-GAN(W). Since combining the word-level and sentence-level rewards achieves better results than using just one of them, we focus more on the combined reward in the following parts.

In review generation and dialogue generation tasks, it is a widely debated question how well the BLEU score against a single reference can reflect the quality of the generated text~\citep{DBLP:conf/emnlp/LiuLSNCP16}. Thus, although the proposed model achieves better BLEU scores compared with baselines, we omit the detailed comparisons in terms of BLEU for space.

%In all, the proposed method overpasses all the existing methods, and enjoys a comfortable margin over SeqGAN. It demonstrates that rewards based on a language model are more effective in learning the distribution of the real data, and also promote diversity in the generated text.

%Combining word-level and sentence-level rewards achieves the better results than using just one of them. Thus, we only evaluate the text generated with the combined reward in human evaluation.

%It can be seen that DP-GAN (S) improves a lot in generating more tokens and DP-GAN (W) has ability to produce much more distinct n-grams.  
%The low the number of distinct n-grams in DP-GAN (S) is because  

%It can be noted that low frequency of Distinct-1, Distinct-2 and Distinct-3 in DP-GAN (S) is because that the proposed model generates more words than MLE. The improvement of total number of generated n-gram is greater than that of distinct n-gram. With the help of word-level reward which substantially improves the number of distinct n-grams, combined DP-GAN  

% Combining word-level and sentence-level rewards achieves the better results than using just one of them. Thus, we only evaluate the text generated with the combined reward in human evaluation.

 %%%%%%%%%%%%%%%%%%%%%%%%%%%%%%%%%%%%%%%%%%%%%%%%%%%%%%%

\begin{figure*}[t] 
\centering
\includegraphics[width = \textwidth]{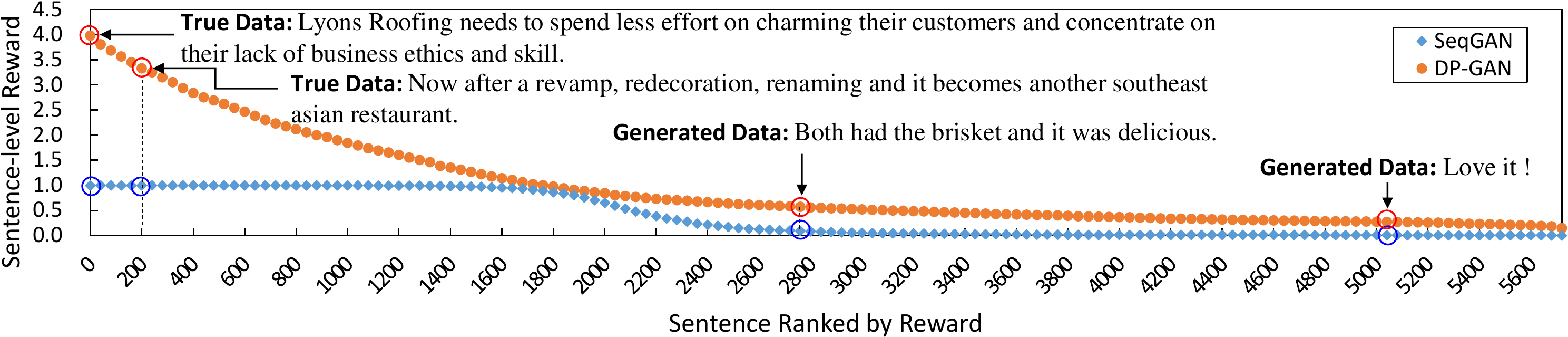} 
\caption{Distribution of rewards between SeqGAN and DP-GAN. The upper two sentences are sampled from the real-world data and the lower two sentences are sampled from  the generated data. It is important to note that the sentence-level reward of DP-GAN is averaged word-level reward and a long sentence does not indicate a high score. As we can see, the reward distribution of SeqGAN saturates and cannot distinguish the novelty of the text accurately. In contrast, DP-GAN has a strong ability of resisting reward saturation and can give more precise reward for text in terms of novelty.}
\label{distribution} 

\end{figure*} 
%%%%%%%%%%%%%%%%%%%%%%%%%%%%%%%%%%%%%%%%%%%%%%%%%%%%%%%%

%%%%%%%%%%%%%%%%%%%%%%%%%%%%%%%%%%%%%%%%%%%%%%%%%%%%%%%%%%%
\begin{figure}[t]
\centering
\includegraphics[width = 0.95\linewidth]{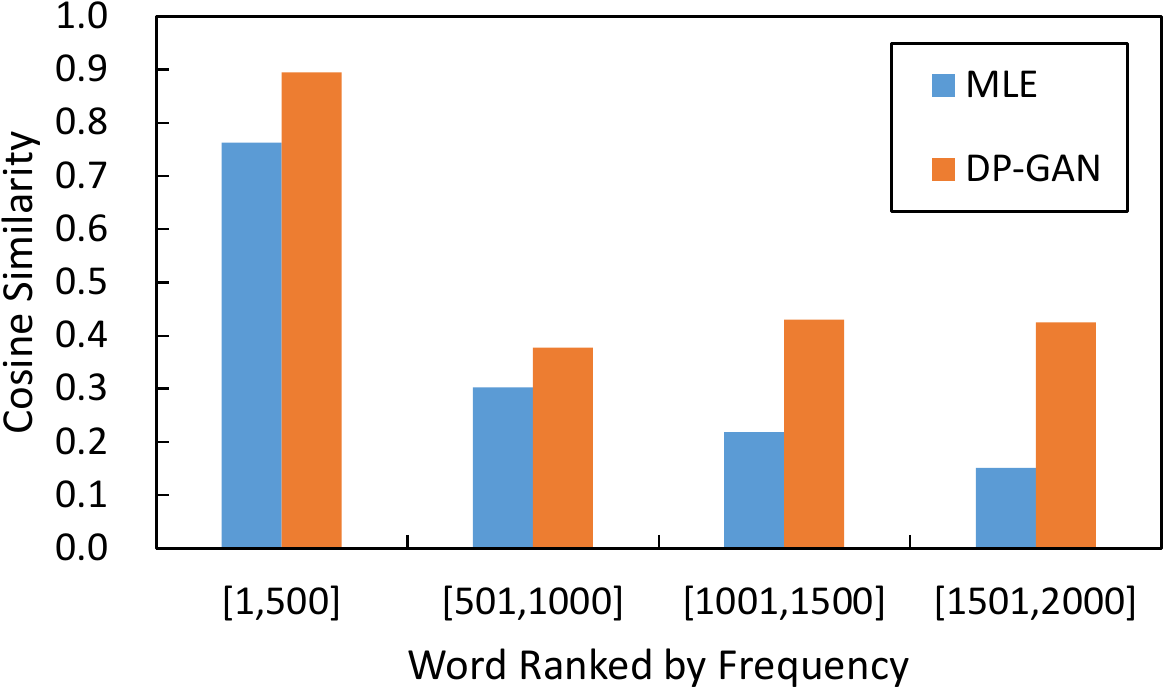} 
\caption{Cosine similarity between the real-world data distribution and the generated data distributions of various models. For example, the first column represents the cosine similarity on top 500 words with the highest frequencies in real-world data. As we can see, the generated data distribution of DP-GAN is closer to the real-world data distribution, especially considering the words of low frequency.}
\label{true}

\end{figure} 
%%%%%%%%%%%%%%%%%%%%%%%%%%%%%%%%%%%%%%%%%%%%%%%%%%%%%%%%%%%%%%%%%

\subsubsection{Human Evaluation}

%We randomly pick 200 items for human evaluation from the test set.
%
We conduct a human evaluation on the test set. For all tasks, we randomly extract 200 samples from the test sets. Each item contains the input text and the text generated by the different systems. The items are distributed to three annotators  who have no knowledge about which system the text is from. Following the work of~\citet{DBLP:conf/emnlp/LiMSJRJ17}, we require them to rank the generated text considering relevance, diversity, and fluency. It is important to note that all the annotators have linguistic background. Relevance means that how likely the generated text is related to the input text. Diversity means that how much the generated text provides specific information, rather than ``dull'' and repeated information. Fluency means that how likely the generated text is produced by human. \textsl{All} represents the ranking given by annotators based on a comprehensive consideration of all human evaluation metrics. 
 The results of human evaluation are shown in Table~\ref{human}.  It needs to be mentioned that in the special case that several pieces of generated text are exactly the same, they are given the same ranking. The inter-annotator agreement is satisfactory considering the difficulty in the human evaluation. The Pearson's correlation coefficient is 0.76 on diversity, 0.59 on fluency and 0.27 on relevance, with $p<0.0001$.  The p-value is all below 0.001, indicating high correlation and agreement. Table~\ref{human} shows that DP-GAN brings the largest improvement in terms of diversity and relevance while scarcely reducing fluency. 
First, DP-GAN significantly outperforms baselines in term of diversity. Second, DP-GAN achieves the best performance in terms of relevance. Diverse text generated by DP-GAN brings more specific expressions, increasing the performance of relevance. For example, given input sentence\textsl{``food is good"}, the relevance of \textsl{``love it, beef is the best"} is obviously better than that of \textsl{``love it"}. Third, the fluency of DP-GAN is slightly worse than that of baselines. It is mainly due to the fact that the text generated by DP-GAN is much longer. From Table~\ref{tab:auto-div}, we can see that the number of words generated by DP-GAN is almost three times than that of baselines in the Yelp dataset. %Generating much longer text while scarcely reducing fluency reflects the strong ability of DP-GAN to generate diverse and good-quality text. 
%this table shows that DP-GAN brings the large improvement in terms of diversity and relevance while scarcely reducing fluency
%Generating much longer text almost without reducing fluency reflects the strong ability of DP-GAN to generate good-quality and diverse text.

%We think that it is an advantage to create much longer text almost without reducing fluency.

%In all, DP-GAN brings the large improvement in terms of diversity and relevance while scarcely reducing fluency.
%It can be concluded that the quality of the text generated by DP-GAN is much better, which is also proved in the fifth column of Table~\ref{human}.
%In all, the results of human performance show that DP-GAN can significantly generate more diverse text almost without reducing fluency. 

%%%%%%%%%%%%%%%%%%%%%%%%%%%%%%%%%%%%%%%%%%%%%%%%%%%%%%%%%%%%%%%
\begin{table}[t]
\small
\centering
    \begin{tabular}{p{0.95\linewidth}}
    \hline
    \textbf{Input}: \textsl{One of my favorite places to eat.}\\
    %\hline
    \textbf{MLE}:  \textsl{Service is great.}    \\  
    \textbf{PG-BLEU}: \textsl{Service is always good.}\\
    \textbf{SeqGAN}: \textsl{Love the chicken and waffles. Service is always great.}\\
    \textbf{DP-GAN}:  \textsl{Love the fact that they have a large selection of food. Service is always great and the food is always fresh. I've been to this place a few times and have never been disappointed.}\\
    
%     %\toprule[2pt]
    \hline
    \textbf{Input}: \textsl{Excellent service, clean, nice ambience.}\\
    %\hline
    \textbf{MLE}:  \textsl{Food is very good.}   \\       
    \textbf{PG-BLEU}: \textsl{Food is very good.}\\
    \textbf{SeqGAN}: \textsl{Food is very good.}\\
    \textbf{DP-GAN}: \textsl{I had the chicken and beef and it was delicious. Food is good and the service is great. I would recommend this place to anyone who wants to eat here.}\\
    
    %\toprule[2pt]
%     \hline
%     \textbf{Input}: Lots of options for dimsum and they still have the carts.\\
%     \textbf{MLE}: service was great.\\
%     \textbf{PG-BLEU}: Service was good.\\
%     \textbf{SeqGAN}: service was good.\\
%     \textbf{DP-GAN}: I had the chicken and waffles, and the chicken was very tender and juicy. Service was good, and the food was good.\\
    
    %\toprule[2pt]
     \hline
    \textbf{Input}: \textsl{I brought friends -- seven of them -- who'd never been to a Bar Louie before.}\\
    %\hline
    \textbf{MLE}: \textsl{Food was mediocre, but the service was great. I will not be returning.}\\
    \textbf{PG-BLEU}: \textsl{Food was terrible. Wouldn't recommend this place to anyone}.\\
    \textbf{SeqGAN}: \textsl{Food was just okay. I wouldn't recommend this place.}\\
    \textbf{DP-GAN}:  \textsl{Ordered a chicken salad with a side of fries, and it was just ok. Food was good, but the service was slow. I had to ask for a refill on my drink, and the bartender was rude. I wouldn't recommend this place to anyone. }\\
    
    %\toprule[2pt]
    \hline
    \textbf{Input}: \textsl{I don't think it's too spicy, so i add a little hot curry spice to the sauce.}\\
    %\hline
    \textbf{MLE}:  \textsl{It's great for cooking. I have to say, I'm not sure if it is the same.}  \\
    \textbf{PG-BLEU}: \textsl{Love it!} \\
    \textbf{SeqGAN}: \textsl{Love it! Love this stuff. I have used it for years and it is very good.}\\
    \textbf{DP-GAN}:\textsl{I've tried many different brands of hot sauces but this one is the best. I've also used it in soups and stews. I also like that it is organic and has a great flavor.}\\
    \hline  
    \end{tabular}
    
    \caption{ Examples generated by DP-GAN and the state-of-the-art models on the Yelp dataset.  It can be found that the text generated by baselines is more generic and repeated, while our model generates text with more specific details and higher diversity. }
   
    \label{samplecases}

\end{table}
%%%%%%%%%%%%%%%%%%%%%%%%%%%%%%%%%%%%%%%%%%%%%%%%%%%%%%
\subsubsection{Analysis: Why It Works}
In this section, we provide detailed analysis to see why our proposed method works better.

In Figure \ref{distribution}, we demonstrate the reward distributions of our model and SeqGAN. %The horizontal axis represents index of samples and the vertical axis represents the value of sentence-level reward. 
%Therefore, the generator does not take diversity into account when generating text, and performs badly in terms of diversity.
%, as the classifier-based discriminator only evaluates whether the text is real
It can be seen that the reward of SeqGAN cannot reflect the novelty of text accurately. 
First, when the novelty of text is relatively high, the reward given by SeqGAN saturates and cannot tell the difference between high-novelty text. Even with different degrees of novelty, the two high-novelty sentences sampled from the true data in the figure get nearly the same reward in SeqGAN. Second, most generated text receives reward around zero because of the high accuracy of classifier. It is hard for such reward to distinguish the difference between low-novelty text. For example, as shown in the figure, ``\textsl{Both had the brisket and it was delicious}'' is much more informative than ``\textsl{Love it!}''. The discriminator of SeqGAN gives them practically the same reward, while the proposed discriminator can better distinguish the two sentences in terms of novelty. In fact, the classifier in SeqGAN trained for 10 epochs can reach very high accuracy, that is, 98.35\% and 99.63\% for Yelp and Amazon, respectively. If the accuracy of classifier is too high, the classifier cannot give reasonable reward to the generator for generating real and diverse text~\citep{DBLP:journals/corr/ArjovskyCB17}.

%Because in reinforcement learning, the generator improves itself by the rewards, if all the rewards are low, no substantial improvements can be made. 

%Hence, the generator is discouraged from generating novel sentences.

%, and tends to repeat the safe expressions, which it learns in pretraining
%The discriminator distinguishes the text more precisely in terms of diversity.
In contrast, the language-model based reward given by DP-GAN better reflect the novelty of the text. The novel text is given high reward that does not saturate. The generated data, which can be less novel, is given relatively low but nonzero reward that can encourage the generator to generate diverse expressions. The refined reward leads to more efficient training, thus resulting in better performance. 
%the proposed method is in line with the work of WGAN. T
%WGAN propose to use Wasserstein distance instead of JSD, which provides clear gradients on all parts of the space.
%we control the gradients using reinforcement learning, and the reward is chosen so that the gradients are sensible to the quality of the text. T

%From another view, the training loss for the binary-classifier based GAN, including SeqGAN, can be regarded as Jensen-Shannon divergence. However, if two distributions are mostly independent, JSD fails to describe the difference between the two distributions and saturates, resulting in vanishing gradients. The vanishing gradients make it hard to encourage the model to produce diverse text. In this work, the language-model based discriminator better differentiates the text, and gives better gradients than the binary-classifier based discriminator, as shown in Figure~\ref{distribution}.

We also compare the cosine similarity between the real-world data distribution and the generated data distributions of various models. Figure~\ref{true} shows the results. We calculate the cosine distance between two vectors, where each element is the frequency of a word indexed by its rank in real-world data. For example, the first element in the vector means the frequency of the word that ranks first in  real-world data. The word frequency vector is divided  into 4 vectors to show the similarity of words of different frequencies. The distribution of words are more similar when they occur more frequently in  real-world data. As DP-GAN promotes diversity, words of low frequency in real-world data are better learned and the similarity is much better than that of MLE. In all, the generated data distribution of DP-GAN is closer to the real-world data distribution in all intervals, especially considering the words of low frequency.

Table \ref{samplecases} presents the examples generated by different models on the Yelp dataset. It can be found that the text generated by MLE is more generic and repeated, while  PG-BLEU and SeqGAN do not perform obviously better than MLE. Moreover, it can be clearly seen that our model generates text with more specific details and higher diversity.

\section{Conclusions}

%It is often observed that the sequence-to-sequence model suffers from the problem of encouraging the model to repeatedly generate simple and generic output. 
In this paper, we propose a new model, called DP-GAN, to promote the diversity of the generated text. DP-GAN assigns low reward for repeated text and high reward for novel and fluent text, encouraging the generator to produce novel and diverse text. We evaluate DP-GAN on two tasks and the findings are concluded as follows: First, the proposed method substantially outperforms the baseline methods in automatic and human evaluations. It shows that DP-GAN is capable of producing more diverse and informative text. Second, the proposed discriminator can better distinguish novel text from repeated text with the saturation problem compared without traditional classifier-based discriminators. Third, with the improvement of diversity, the generated data distribution of DP-GAN is closer to the real-world data distribution compared with that of MLE.

\section*{Acknowledgements}

This work was supported in part by National Natural Science Foundation of China (No. 61673028). We thank Wei Li and Bingzhen Wei for providing the thoughtful suggestions.   Xu Sun is the corresponding author of this paper.

\end{CJK}

\bibliography{emnlp2018}

\begin{thebibliography}{35}
\expandafter\ifx\csname natexlab\endcsname\relax\def\natexlab#1{#1}\fi

\bibitem[{Arjovsky et~al.(2017)Arjovsky, Chintala, and
  Bottou}]{DBLP:journals/corr/ArjovskyCB17}
Mart{\'{\i}}n Arjovsky, Soumith Chintala, and L{\'{e}}on Bottou. 2017.
\newblock Wasserstein generative adversarial networks.
\newblock In \emph{{ICML} 2017}, pages 214--223.

\bibitem[{Bahdanau et~al.(2017)Bahdanau, Brakel, Xu, Goyal, Lowe, Pineau,
  Courville, and Bengio}]{DBLP:journals/corr/BahdanauBXGLPCB16}
Dzmitry Bahdanau, Philemon Brakel, Kelvin Xu, Anirudh Goyal, Ryan Lowe, Joelle
  Pineau, Aaron~C. Courville, and Yoshua Bengio. 2017.
\newblock An actor-critic algorithm for sequence prediction.
\newblock In \emph{ICLR 2017}.

\bibitem[{Bahdanau et~al.(2014)Bahdanau, Cho, and
  Bengio}]{DBLP:journals/corr/BahdanauCB14}
Dzmitry Bahdanau, Kyunghyun Cho, and Yoshua Bengio. 2014.
\newblock Neural machine translation by jointly learning to align and
  translate.
\newblock In \emph{{ICLR} 2014}.

\bibitem[{Bengio et~al.(2015)Bengio, Vinyals, Jaitly, and
  Shazeer}]{DBLP:conf/nips/BengioVJS15}
Samy Bengio, Oriol Vinyals, Navdeep Jaitly, and Noam Shazeer. 2015.
\newblock Scheduled sampling for sequence prediction with recurrent neural
  networks.
\newblock In \emph{NIPS 2015}, pages 1171--1179.

\bibitem[{Berthelot et~al.(2017)Berthelot, Schumm, and
  Metz}]{DBLP:journals/corr/BerthelotSM17}
David Berthelot, Tom Schumm, and Luke Metz. 2017.
\newblock {BEGAN:} boundary equilibrium generative adversarial networks.
\newblock \emph{CoRR}, abs/1703.10717.

\bibitem[{Chen et~al.(2016)Chen, Duan, Houthooft, Schulman, Sutskever, and
  Abbeel}]{DBLP:conf/nips/ChenCDHSSA16}
Xi~Chen, Yan Duan, Rein Houthooft, John Schulman, Ilya Sutskever, and Pieter
  Abbeel. 2016.
\newblock Infogan: Interpretable representation learning by information
  maximizing generative adversarial nets.
\newblock In \emph{NIPS 2016}, pages 2172--2180.

\bibitem[{Cho et~al.(2014)Cho, van Merrienboer, G{\"{u}}l{\c{c}}ehre, Bahdanau,
  Bougares, Schwenk, and Bengio}]{DBLP:conf/emnlp/ChoMGBBSB14}
Kyunghyun Cho, Bart van Merrienboer, {\c{C}}aglar G{\"{u}}l{\c{c}}ehre, Dzmitry
  Bahdanau, Fethi Bougares, Holger Schwenk, and Yoshua Bengio. 2014.
\newblock Learning phrase representations using {RNN} encoder-decoder for
  statistical machine translation.
\newblock In \emph{{EMNLP} 2014}, pages 1724--1734.

\bibitem[{Denton et~al.(2015)Denton, Chintala, Szlam, and
  Fergus}]{DBLP:conf/nips/DentonCSF15}
Emily~L. Denton, Soumith Chintala, Arthur Szlam, and Rob Fergus. 2015.
\newblock Deep generative image models using a laplacian pyramid of adversarial
  networks.
\newblock In \emph{NIPS 2015}, pages 1486--1494.

\bibitem[{Duchi et~al.(2011)Duchi, Hazan, and
  Singer}]{DBLP:journals/jmlr/DuchiHS11}
John~C. Duchi, Elad Hazan, and Yoram Singer. 2011.
\newblock Adaptive subgradient methods for online learning and stochastic
  optimization.
\newblock \emph{Journal of Machine Learning Research}, 12:2121--2159.

\bibitem[{Goodfellow et~al.(2014)Goodfellow, Pouget{-}Abadie, Mirza, Xu,
  Warde{-}Farley, Ozair, Courville, and
  Bengio}]{DBLP:conf/nips/GoodfellowPMXWOCB14}
Ian~J. Goodfellow, Jean Pouget{-}Abadie, Mehdi Mirza, Bing Xu, David
  Warde{-}Farley, Sherjil Ozair, Aaron~C. Courville, and Yoshua Bengio. 2014.
\newblock Generative adversarial nets.
\newblock In \emph{NIPS 2014}, pages 2672--2680.

\bibitem[{Gulrajani et~al.(2017)Gulrajani, Ahmed, Arjovsky, Dumoulin, and
  Courville}]{DBLP:conf/nips/GulrajaniAADC17}
Ishaan Gulrajani, Faruk Ahmed, Mart{\'{\i}}n Arjovsky, Vincent Dumoulin, and
  Aaron~C. Courville. 2017.
\newblock Improved training of wasserstein gans.
\newblock In \emph{NIPS 2017}, pages 5769--5779.

\bibitem[{Guu et~al.(2017)Guu, Hashimoto, Oren, and
  Liang}]{DBLP:journals/corr/abs-1709-08878}
Kelvin Guu, Tatsunori~B. Hashimoto, Yonatan Oren, and Percy Liang. 2017.
\newblock Generating sentences by editing prototypes.
\newblock \emph{CoRR}, abs/1709.08878.

\bibitem[{Li et~al.(2016)Li, Galley, Brockett, Gao, and
  Dolan}]{DBLP:conf/naacl/LiGBGD16}
Jiwei Li, Michel Galley, Chris Brockett, Jianfeng Gao, and Bill Dolan. 2016.
\newblock A diversity-promoting objective function for neural conversation
  models.
\newblock In \emph{{NAACL} 2016}, pages 110--119.

\bibitem[{Li et~al.(2015)Li, Luong, and Jurafsky}]{DBLP:conf/acl/LiLJ15}
Jiwei Li, Minh{-}Thang Luong, and Dan Jurafsky. 2015.
\newblock A hierarchical neural autoencoder for paragraphs and documents.
\newblock In \emph{{ACL} 2015}, pages 1106--1115.

\bibitem[{Li et~al.(2017)Li, Monroe, Shi, Jean, Ritter, and
  Jurafsky}]{DBLP:conf/emnlp/LiMSJRJ17}
Jiwei Li, Will Monroe, Tianlin Shi, S{\'{e}}bastien Jean, Alan Ritter, and Dan
  Jurafsky. 2017.
\newblock Adversarial learning for neural dialogue generation.
\newblock In \emph{{EMNLP} 2017}, pages 2157--2169.

\bibitem[{Lin et~al.(2018)Lin, Sun, Ma, and
  Su}]{DBLP:journals/corr/abs-1805-03989}
Junyang Lin, Xu~Sun, Shuming Ma, and Qi~Su. 2018.
\newblock Global encoding for abstractive summarization.
\newblock \emph{CoRR}, abs/1805.03989.

\bibitem[{Liu et~al.(2016)Liu, Lowe, Serban, Noseworthy, Charlin, and
  Pineau}]{DBLP:conf/emnlp/LiuLSNCP16}
Chia{-}Wei Liu, Ryan Lowe, Iulian Serban, Michael Noseworthy, Laurent Charlin,
  and Joelle Pineau. 2016.
\newblock How {NOT} to evaluate your dialogue system: An empirical study of
  unsupervised evaluation metrics for dialogue response generation.
\newblock In \emph{{EMNLP} 2016}, pages 2122--2132.

\bibitem[{Liu et~al.(2017)Liu, Wang, Sha, Chang, and
  Sui}]{DBLP:journals/corr/abs-1711-09724}
Tianyu Liu, Kexiang Wang, Lei Sha, Baobao Chang, and Zhifang Sui. 2017.
\newblock Table-to-text generation by structure-aware seq2seq learning.
\newblock \emph{CoRR}, abs/1711.09724.

\bibitem[{Luo et~al.(2018)Luo, Xu, Lin, Zeng, and Sun}]{jingjingxuemnlp18-02}
Liangchen Luo, Jingjing Xu, Junyang Lin, Qi~Zeng, and Xu~Sun. 2018.
\newblock An auto-encoder matching model for learning utterance-level semantic
  dependency in dialogue generation.
\newblock In \emph{{EMNLP}, 2018}.

\bibitem[{Luong et~al.(2015)Luong, Pham, and
  Manning}]{DBLP:conf/emnlp/LuongPM15}
Thang Luong, Hieu Pham, and Christopher~D. Manning. 2015.
\newblock Effective approaches to attention-based neural machine translation.
\newblock In \emph{{EMNLP} 2015}, pages 1412--1421.

\bibitem[{Ma et~al.(2018{\natexlab{a}})Ma, Sun, Lin, and
  Wang}]{DBLP:journals/corr/abs-1805-04869}
Shuming Ma, Xu~Sun, Junyang Lin, and Houfeng Wang. 2018{\natexlab{a}}.
\newblock Autoencoder as assistant supervisor: Improving text representation
  for chinese social media text summarization.
\newblock \emph{CoRR}, abs/1805.04869.

\bibitem[{Ma et~al.(2018{\natexlab{b}})Ma, Sun, Wang, and
  Lin}]{DBLP:journals/corr/abs-1805-04871}
Shuming Ma, Xu~Sun, Yizhong Wang, and Junyang Lin. 2018{\natexlab{b}}.
\newblock Bag-of-words as target for neural machine translation.
\newblock \emph{CoRR}, abs/1805.04871.

\bibitem[{McAuley and Leskovec(2013)}]{McAuley2013}
Julian~John McAuley and Jure Leskovec. 2013.
\newblock From amateurs to connoisseurs: modeling the evolution of user
  expertise through online reviews.
\newblock In \emph{{WWW} 2013}, pages 897--908.

\bibitem[{Nowozin et~al.(2016)Nowozin, Cseke, and Tomioka}]{f-GAN}
Sebastian Nowozin, Botond Cseke, and Ryota Tomioka. 2016.
\newblock f-gan: Training generative neural samplers using variational
  divergence minimization.
\newblock In \emph{NIPS 2016}, pages 271--279.

\bibitem[{Radford et~al.(2015)Radford, Metz, and
  Chintala}]{DBLP:journals/corr/RadfordMC15}
Alec Radford, Luke Metz, and Soumith Chintala. 2015.
\newblock Unsupervised representation learning with deep convolutional
  generative adversarial networks.
\newblock \emph{CoRR}, abs/1511.06434.

\bibitem[{Ranzato et~al.(2016)Ranzato, Chopra, Auli, and
  Zaremba}]{DBLP:journals/corr/RanzatoCAZ15}
Marc'Aurelio Ranzato, Sumit Chopra, Michael Auli, and Wojciech Zaremba. 2016.
\newblock Sequence level training with recurrent neural networks.
\newblock In \emph{ICLR 2016}.

\bibitem[{Salimans et~al.(2016)Salimans, Goodfellow, Zaremba, Cheung, Radford,
  and Chen}]{DBLP:conf/nips/SalimansGZCRCC16}
Tim Salimans, Ian~J. Goodfellow, Wojciech Zaremba, Vicki Cheung, Alec Radford,
  and Xi~Chen. 2016.
\newblock Improved techniques for training gans.
\newblock In \emph{NIPS 2016}, pages 2226--2234.

\bibitem[{Shao et~al.(2017)Shao, Gouws, Britz, Goldie, Strope, and
  Kurzweil}]{DBLP:conf/emnlp/ShaoGBGSK17}
Yuanlong Shao, Stephan Gouws, Denny Britz, Anna Goldie, Brian Strope, and Ray
  Kurzweil. 2017.
\newblock Generating high-quality and informative conversation responses with
  sequence-to-sequence models.
\newblock In \emph{{EMNLP} 2017}, pages 2210--2219.

\bibitem[{Sutskever et~al.(2014)Sutskever, Vinyals, and
  Le}]{DBLP:conf/nips/SutskeverVL14}
Ilya Sutskever, Oriol Vinyals, and Quoc~V. Le. 2014.
\newblock Sequence to sequence learning with neural networks.
\newblock In \emph{NIPS 2014}, pages 3104--3112.

\bibitem[{Sutton et~al.(1999)Sutton, McAllester, Singh, and
  Mansour}]{DBLP:conf/nips/SuttonMSM99}
Richard~S. Sutton, David~A. McAllester, Satinder~P. Singh, and Yishay Mansour.
  1999.
\newblock Policy gradient methods for reinforcement learning with function
  approximation.
\newblock In \emph{NIPS 1999}, pages 1057--1063.

\bibitem[{Williams(1992)}]{DBLP:journals/ml/Williams92}
Ronald~J. Williams. 1992.
\newblock Simple statistical gradient-following algorithms for connectionist
  reinforcement learning.
\newblock \emph{Machine Learning}, 8:229--256.

\bibitem[{Xu et~al.(2018{\natexlab{a}})Xu, Sun, Zeng, Ren, Zhang, Wang, and
  Li}]{unpaired-sentiment-translation}
Jingjing Xu, Xu~Sun, Qi~Zeng, Xuancheng Ren, Xiaodong Zhang, Houfeng Wang, and
  Wenjie Li. 2018{\natexlab{a}}.
\newblock Unpaired sentiment-to-sentiment translation: A cycled reinforcement
  learning approach.
\newblock In \emph{{ACL}, 2018}.

\bibitem[{Xu et~al.(2018{\natexlab{b}})Xu, Zhang, Zeng, Ren, Cai, and
  Sun}]{jingjingxuemnlp18-01}
Jingjing Xu, Yi~Zhang, Qi~Zeng, Xuancheng Ren, Xiaoyan Cai, and Xu~Sun.
  2018{\natexlab{b}}.
\newblock A skeleton-based model for promoting coherence among sentences in
  narrative story generation.
\newblock In \emph{{EMNLP}, 2018}.

\bibitem[{Yu et~al.(2017)Yu, Zhang, Wang, and Yu}]{DBLP:conf/aaai/YuZWY17}
Lantao Yu, Weinan Zhang, Jun Wang, and Yong Yu. 2017.
\newblock Seqgan: Sequence generative adversarial nets with policy gradient.
\newblock In \emph{{AAAI} 2017}, pages 2852--2858.

\bibitem[{Zhao et~al.(2017)Zhao, Mathieu, and
  LeCun}]{DBLP:journals/corr/ZhaoML16}
Junbo~Jake Zhao, Micha{\"{e}}l Mathieu, and Yann LeCun. 2017.
\newblock Energy-based generative adversarial network.
\newblock In \emph{ICLR 2017}.

\end{thebibliography}
\bibliographystyle{acl_natbib_nourl}
\end{document}